\documentclass[conference]{IEEEtran}
\IEEEoverridecommandlockouts
\usepackage{cite}
\usepackage{amsmath,amssymb,amsfonts}
\usepackage{algorithmic}
\usepackage{graphicx}
\usepackage{subfig} 
\usepackage{textcomp}
\usepackage{xcolor}
\usepackage{adjustbox}
\usepackage{float}

\usepackage{booktabs} % For better looking tables
\usepackage{enumitem} % For customizing list formatting

\def\BibTeX{{\rm B\kern-.05em{\sc i\kern-.025em b}\kern-.08em
    T\kern-.1667em\lower.7ex\hbox{E}\kern-.125emX}}
\begin{document}

\newcommand{\confName}{ICCIT }

\title{BornoViT: A Novel Efficient Vision Transformer for Bengali Handwritten Basic Characters Classification  
}

%\author{(Anonymous \confName  submission)}

\author{ 
    \IEEEauthorblockN{
    Rafi Hassan Chowdhury\textsuperscript{1}, 
    Naimul Haque\textsuperscript{1}, 
    Kaniz Fatiha\textsuperscript{1}\\}

    \IEEEauthorblockA{
        \textsuperscript{1}Department of Computer Science and Engineering, Islamic University of Technology, Gazipur 1704, Bangladesh}

    \IEEEauthorblockA{\{rafihassan, naimulhaque, kanizfatiha\}@iut-dhaka.edu
    } 
}

%%%%%%% ADDING CONFERENCE info in TOP-LEFT Corner %%%%%%%%%%
%  Source: https://tex.stackexchange.com/questions/561793/how-to-get-ieee-conference-template-to-show-conference-name-in-header

\makeatletter
\let\old@ps@IEEEtitlepagestyle\ps@IEEEtitlepagestyle
\def\confheader#1{%
    % for the first page
    \def\ps@IEEEtitlepagestyle{%
        \old@ps@IEEEtitlepagestyle%
        \def\@oddhead{\strut\hfill#1\hfill\strut}%
        \def\@evenhead{\strut\hfill#1\hfill\strut}%
    }%
    \ps@headings%
}
\makeatother
\confheader{
        \parbox{20cm}{2024 27th International Conference on Computer and Information Technology (ICCIT)\\
        20-22 December 2024, Cox’s Bazar, Bangladesh}
}

\IEEEpubid{
\begin{minipage}[t]{\textwidth}\ \\[10pt]
      \small{979-8-3315-1909-4/24/\$31.00 \copyright2024 IEEE }
\end{minipage}
}

\maketitle

\begin{abstract}
Handwritten Character Classification in Bengali script is a significant challenge due to the complexity and variability of the characters. The models which are used often, to classify characters are computationally expensive and data-hungry which are not suitable for the resource limited languages such as Bengali. In this experiment, we proposed a novel, efficient and lightweight Vision Transformer model that can classify Bengali handwritten basic characters \& digits effectively, addressing some of the shortcomings of traditional methods. The proposed solution utilizes deep convolutional neural network (DCNN) in a more simplified mannar than in traditional DCNN with the aim of lowering the computational burden. With only 0.65 million parameters, 0.62 MB of model size and 0.16 GFLOPs, our model BornoViT is much lighter than the current state-of-the-art models,  making our model more favorable for resource limited environment, which is quite necessary for Bengali handwritten character classification. Our model BornoViT was evaluated on BanglaLekha Isolated dataset achieving an accuracy of 95.77\%, which significantly outperforms current state-of-the-art models in terms of efficiency.  Furthermore, the model was evaluated using our own captured dataset, \textit{Bornomala}, consisting of around 222 samples from different age groups, achieving accuracy of 91.51\%.

\end{abstract}


\section{Introduction}

Bangla is the seventh most spoken language in the world, yet Handwritten Character Rcognition (HCR) in Bangla scripts still requires further research because of its complex nature and characteristic. The Bangla letters have complex and distinct patterns, making it challenging for models to achieve high accuracy when classifying a wide range of different handwriting patterns.
Traditional machine learning techniques that rely on engineered descriptors often fall short in addressing these complexities. Moreover, most of existing models are computationally expensive, which are not suitable for low end devices. 

Additionally, template matching methods are not suitable because of the feature variations, writing pattern of different people and inter-class similarity in Bengali handwritten characters. While convolutional Neural Networks (CNNs) in deep learning models are able to overcome certain challenges, but they fall behind in terms of identifying spatial patterns effectively. This is where Vision Transformers (ViTs) become valuable, as they utilize attention mechanisms to learn from global spatial features. As a result, this approach is useful to classify salient features and patterns in context, without the need of expensive computations. Such method is particularly useful for tasks like Bengali handwritten character recognition.

Our work introduces a novel, lightweight model called
BornoViT, proposed specifically for the efficient classification of
Bengali handwritten basic characters. Our model, based on simplified Vision Transformer (ViT) architecture, focusing on reducing computational complexity without compromising classification
accuracy. BornoViT has significantly low parameter count, GFLOPs and model size compared to existing models, while achieving higher accuracy and being less resource-hungry, which makes it suitable for real-time applications in low-end devices. We have evaluated our model on the existing BanglaLekha-Isolated dataset \cite{biswas2017banglalekha}, and it outperformed the current state-of-the-art accuracy in terms of parameter count, model size and FLOPs count.  Along with that, our model performed good on our own dataset, Bornomala. 

\section{Literature Review}

Bangla handwritten character recognition has achieved remarkable progress owing to the implementation of various machine learning methods. VashaNet, developed by Raquib et al (2024) \cite{RAQUIB2024100568} where the 26-skeleton deep Convolutional Neural Network model is aimed at looking for Bengali handwritten characters. VashaNet recorded an accuracy of 94.60\% on a primary data set and 94.43\% accuracy in the CMATERdb data set. 

Previously, in their work, Chauhan et al. (2024) \cite{chauhan2024hcr} introduced HCR-Net, a deep learning-based script-independent handwritten character recognition network, aimed at improving recognition accuracy across various languages through transfer learning and image augmentation techniques which achieved performance improvements of up to 11\% over existing models, making it a computationally efficient solution for handwritten character recognition, especially on smaller datasets. 

Similarly, in the context of Bangla character recognition, Opu et al. (2021) \cite{opu2024handwritten} proposed a lightweight deep learning model using CNNs which consisted of 74 layers and five blocks, was designed for resource-constrained devices, achieving competitive accuracy with a reduced model size and faster computation times.

Previously, in their work, Sarkhel et al (2016) \cite{SARKHEL2016172} discussed a low-cost method on Bangla character recognition using Support Vector Machines which was tested on the dataset CMATERdb where the model was quite promising from the computational effectiveness perspective but had challenges when operating large and complex datasets.

Further development was reported by Rabby et al. (2020),\cite{inbook} who developed the Borno multiclass CNN based architecture trained on a million images from several datasets where the model reached the impressive level of 92.61 percent of accuracy but it demands high computation resources and so it was difficult to deploy in real-time applications and in settings with low computation resources.

Das et al. (2021) \cite{das2021bangla} explored recognition for Bangla handwriting consonants based on either syllabic or consonantal framing or both, by means of a CNN equiped model and this provides accuracy rates of 93.18\% for vowels and 92.25\% for consonants, thus displaying the capabilities of CNNs against the different portions of Bangla script.

Maitra et al. (2015) \cite{maitra2015cnn} implemented CNN models in the tasks of numeral recognition in various scripts including Bangla where they managed to detect these numerals but this study was insufficient in analyzing the complexity of recognizing the entire Bangla character set \cite{maitra2015cnn}. 

CNN-Bengali\cite{asraful2023handwritten} introduced a simple and efficient CNN framework designed to effectively classify handwritten Bangla characters, including simple characters, compound characters, and numerals where their model demonstrated fast execution, required fewer training epochs, and achieved high validation accuracy across different datasets.

\section{Methodology}
At the beginning of our experiment, we have pre-trained our proposed BornoViT using Ekush dataset \cite{ekush}. After that We have conducted experiment on our Own Dataset \textit{Bornomala} and BanglaLekha-Isolated \cite{biswas2017banglalekha}. We have utilized k-fold cross validation for each dataset, where value of \textit{k} was five. Each time, three folds were used for training, one fold for validation and one for testing split. Validation split was used for early stopping to overcome overfitting of the model. We have trained each fold for at most 100 epochs and our model was converged within 60 to 70 epochs. To stop early we triggered patience if the average validation loss is less than the best validation loss. If patience is triggered for continuous ten times, the training will be stopped.  

\subsection{Datasets}
To conduct this experiment, we have collected our own dataset and also utilized BanglaLekha-Isolated \cite{biswas2017banglalekha} and Ekush \cite{ekush} dataset.
Our dataset includes total 13318 images of 60 classes Bengali Handwritten Characters, which are grouped into three main categories: 11 vowels (Shorborno), 39 consonants (Benjonborno), and 10 Bengali digits. A total of 222 participants took part, involving students at different levels of education, family members and other professionals as well as spanning all age groups. This in itself makes it rich as the handwriting of humans is naturally variant and varies in stroke thickness, character forms, writing style and this anything different perturbs a raid network.

\begin{figure}[htbp] % Changed from [htbp] to [H]
    \centering
    \includegraphics[width=0.45\textwidth]{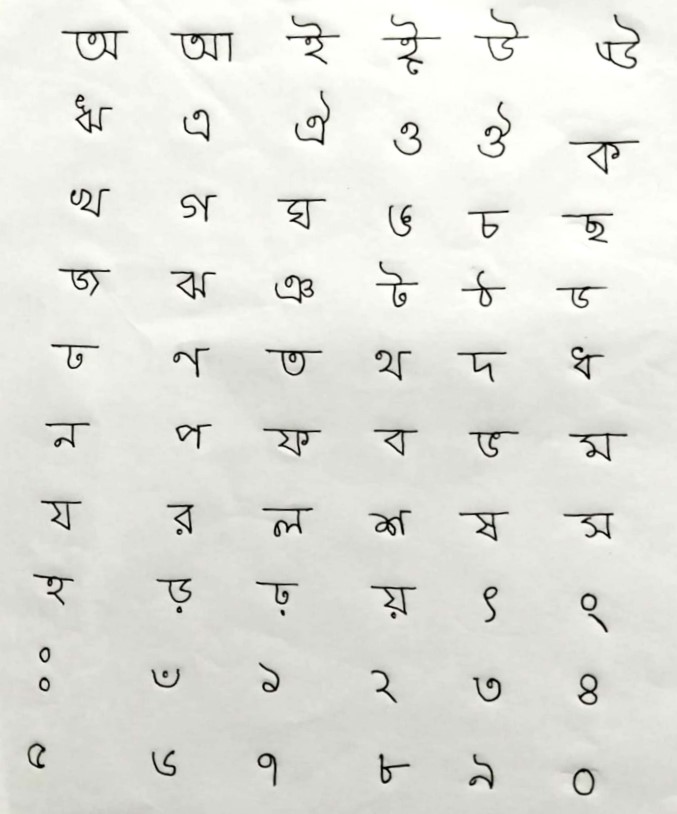}
    \caption{Sample of our Dataset}
    \label{fig:sample_of_our_dataset}
\end{figure}

Every individual was given an A4 sized paper with 10 rows of 6 columns on which to write 60 characters per page. First, we care about type style that including size, weight and stroke, So layout are made very carefully. A high-resolution 800 dpi scan was used to capture the fine details and stroke of characters on all handwritten pages. As the OCR rounds returned each character, these characters were cropped out to make up separate images of characters. The main aim of this data collection is to provide a robust and balanced dataset that contains diverse samples of the handwritten Bengali characters. This variety was necessary to give the model a chance of working well in an enormous number of writing styles, sizes and shapes. Figure \ref{fig:sample_of_our_dataset} shows some samples of our dataset.

\begin{figure}[htbp] % Changed from [htbp] to [H]
    \centering
    \includegraphics[width=0.45\textwidth]{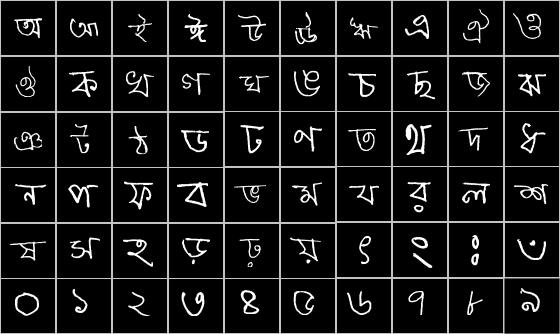}
    \caption{Sample of BanglaLekha-Isolated Dataset}
    \label{fig:sample_of_BanglaLekha-Isolated_dataset}
\end{figure}

In addition to our created dataset Bornomala, BanglaLekha Isolated \cite{biswas2017banglalekha} was utilized in this study. The BanglaLekha-Isolated dataset, created by Biswas et al. \cite{biswas2017banglalekha}, comprises 166,105 squared images, maintaining the aspect ratio of the characters. It includes 84 distinct Bangla characters, which are divided into 50 basic characters, 10 numerals, and 24 compound characters. This dataset is extensively utilized in handwriting recognition tasks and in building OCR systems for the Bangla language. Figure \ref{fig:sample_of_BanglaLekha-Isolated_dataset} shows sample from BanglaLekha-Isolated. A comparison between the three datasets are given in Table. 1.

We have pretrained our BornoViT model by utilizing Ekush Dataset \cite{ekush}. Vision Transformer needs to be pretrained primarily because of their reliance on large datasets for effective learning as they lack the spatial inductive biases presents in their Convolutional Neural Networks (CNNs).

\begin{table}[htbp]
\caption{Comparison of Bangla Handwritten Character Datasets}
\centering
\begin{adjustbox}{width=\linewidth}
\begin{tabular}{ l  c c }
\hline
\textbf{Dataset Name}    & \textbf{Basic Characters} & \textbf{Total Class} \\ \hline

BanglaLekha-Isolated \cite{biswas2017banglalekha}  & 98,950  & 84   \\ 
Ekush  \cite{ekush}  & 154,824   & 122      \\
Bornomala (Ours)   &  13,000   & 60  \\ \hline
\end{tabular}
\end{adjustbox}

\label{tab:datasets}
\end{table}

\subsection{Data Preprocessing}
\begin{figure}[H]
    \centering
    \begin{tabular}{cccc}

        \hspace{-0.3cm}
        
        \subfloat[Original]{\includegraphics[width=0.12\textwidth]{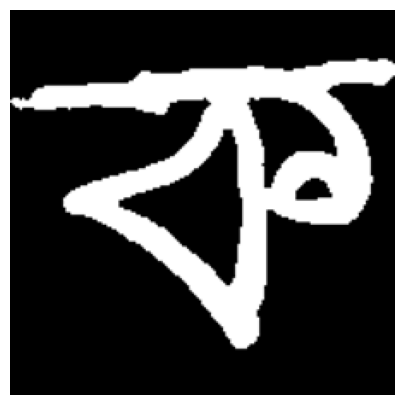}} &

        \hspace{-0.5cm}
        
        \subfloat[Width and Height Shift]{\includegraphics[width=0.12\textwidth]{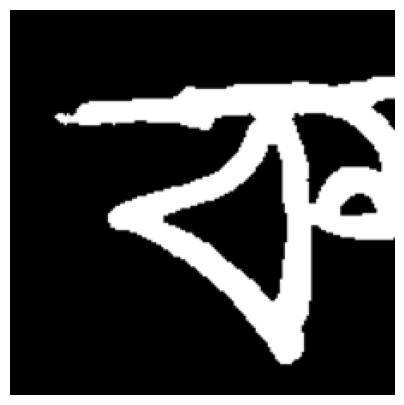}}  &

        \hspace{-0.5cm}
        
        \subfloat[Shearing]{\includegraphics[width=0.12\textwidth]{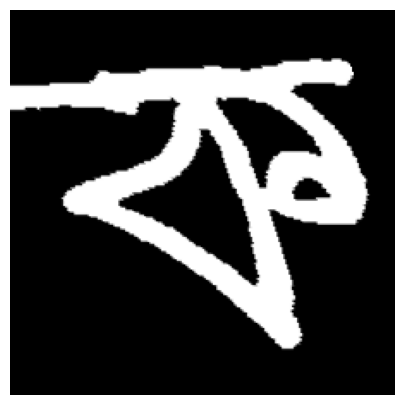}} &

        \hspace{-0.5cm}
        
        \subfloat[Color Jitter]{\includegraphics[width=0.12\textwidth]{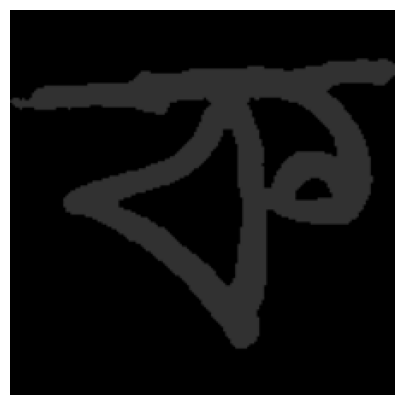}}  \\
    \end{tabular}
    \caption{Sample of augmentation techniques}
    \label{fig:aug_sample}
\end{figure}

We have utilized various Data Preprocessing and Augmentation Techniques. These techniques were applied on runtime, to make the model robustness and generalized for Bengali basic character recognition. As our ViT takes 224x224 pixels input, we transform all input to that size. After that combination of Random Affine and Color Jitter was applied. For Random affine, height and width shift were applied in the range of 0 to 0.1 and shearing with value 20. We applied color jitter with settings that adjust brightness, contrast, and saturation by up to ±20\%, while allowing hue to shift by ±10\%. Figure \ref{fig:aug_sample} show some sample of our augmentation techniques.

\subsection{Model Architecture}

Vision Transformer (ViT) is a deep learning model architected for image classification, inspired from the transformer models used in Natural Language Processing (NLP). It processes input by dividing into non-overlapping patches, after that these patches are flattened and embedded into vectors. These embeddings are fed into transformer layers that apply self-attention mechanisms to capture the global relationships between patches, enabling the model to classify images effectively. Unlike CNNs, ViT eliminates the need for convolutions, relying instead on transformers' ability to model long-range dependencies.

The architecture of our proposed Vision Transformer model is presented in Table \ref{tab:layer_wise_breakdown_of_custom_model} \& \ref{tab:transformer_block_breakdown}. This model consists of a patch embedding layer, positional encoding, and multiple transformer blocks, followed by a classification head.

% Table for Our Proposed Vision Transformer

\begin{table*}[htbp]
    \centering
    \caption{Layer-wise Breakdown of Custom Vision Transformer}
    \begin{tabular}{@{}l p{5.5cm} l l l @{}}
        \toprule
        \textbf{Layer} & \textbf{Description} & \textbf{Input Shape} & \textbf{Output Shape} & \textbf{Parameters} \\ \midrule
        \vspace{0.1cm}
        
        PatchEmbedding & Conv2D layer to split image into patches and project them to embedding dimension & (B, 3, 224, 224) & (B, 196, 128) & 98,432  \\
        \vspace{0.1cm}
        
        Positional Embedding & Learnable positional embedding added to the input patches & (B, 197, 128) & (B, 197, 128) & 25,216 \\
        \vspace{0.1cm}
        
        CLS Token & Learnable CLS token for classification & (1, 1, 128) & (B, 1, 128) & 128   \\
        \vspace{0.1cm}
        
        Dropout & Dropout applied to the patch and position embeddings & (B, 197, 128) & (B, 197, 128) & 0 \\
        \vspace{0.1cm}
        
        Transformer Blocks (x4) & Sequence of 4 Transformer blocks & (B, 197, 128) & (B, 197, 128) & 4x132,096 (Table \ref{tab:transformer_block_breakdown})   \\
        \vspace{0.1cm}
        
        LayerNorm & Normalize embeddings & (B, 197, 128) & (B, 197, 128) & 256  \\
        \vspace{0.1cm}
        
        Linear (Classification) & Final linear layer to classify based on CLS token & (B, 128) & (B, 10) & 1,290 \\ 
        \hline
        
        \textbf{Total} & & & & \textbf{6,53,706} \\ \bottomrule
    \end{tabular}
    \label{tab:layer_wise_breakdown_of_custom_model}
\end{table*}

 % Add vertical space
Each transformer block contains two LayerNorm layers, a multi-head attention mechanism, and an MLP block. Table \ref{tab:transformer_block_breakdown} shows breakdown of the transformer block. 

\begin{table*}[htbp]
    \centering
    \caption{Transformer Block Breakdown}
    \begin{tabular}{@{} l p{5.5cm} l l l@{}}
        \toprule
        \textbf{Sub-layer} & \textbf{Description} & \textbf{Input Shape} & \textbf{Output Shape} & \textbf{Parameters} \\ 
        \midrule
        \vspace{0.1cm}
        
        LayerNorm 1 & Layer normalization & (B, 197, 128) & (B, 197, 128) & 256 \\
        \vspace{0.1cm}
        
        Multi-Head Attention & Self-attention mechanism with 2 attention heads & (B, 197, 128) & (B, 197, 128) & 49,152 (QKV), 16,384 (O) \\
        \vspace{0.1cm}
        
        LayerNorm 2 & Layer normalization & (B, 197, 128) & (B, 197, 128) & 256  \\
        \vspace{0.1cm}
        
        MLP & Two linear layers with GELU and dropout & (B, 197, 128) & (B, 197, 128) & 33,024 (fc1), 33,024 (fc2)  \\ \hline
        
        \textbf{Total} & & & & \textbf{132,096} \\ \bottomrule
    \end{tabular}
    \label{tab:transformer_block_breakdown}
\end{table*}

Our proposed BornoViT architecture follows a patch-based approach for image classification. At the beginning the input image is divided into 16x16 non-overlapping patches, each of which is flattened into a 128-dimensional vector. A prepended learnable classification token (CLS) is added to the patch sequence and a positional embedding is applied to each patch embedding so that it can preserve spatial information.

After that, the patch embeddings, including the CLS token, are fed into four transformer blocks. Each block is made up of a multi-head self-attention mechanism with a subsequent feed-forward network. After each block, layer normalization and residual connections are utilized to prevent overfitting during training.

After the input sequence passes through the transformer blocks, the CLS token is extracted and fed into a linear classifier to make final prediction. These transformer blocks helps the model to learn dependencies between all image patches, making it highly effective for image classification.

Each transformer block has two core components: multi-head self-attention and a multi-layer perceptron (MLP). The self-attention allows the model to focus on different area of the image at once, gathering contextual information from different area. The MLP then fine-tines these representations by utilizing non-linear transformations. This enables the model to understand salient features.

\subsection{Transfer Learning}
Transfer learning allows a model to utilize knowledge gained from one task to another. By starting with a pretrained model in classification tasks, training becomes faster and performance improves. Due to the lack of inductive biases, ViT struggles to generalize on smaller datasets, making transfer learning more effective for such cases. Studies have shown that when ViT is pretrained on large datasets like ImageNet-21k \cite{deng2009imagenet}, and those weights are used for fine-tuning on smaller datasets, it achieves remarkable results, surpassing several image recognition benchmarks\cite{raghu2021vision}. In our experiment, we have trained our BornoViT on Ekush Dataset \cite{ekush} for 100 epochs and then conduct our experiment. As a result, we were able to overcome the lacking of inductive biases.

\subsection{Experimental Setup}

The experiment was conducted on a desktop with a Intel Core i5-13500 processor running at 2.50 GHz. An RTX 3060 GPU was utilized for model training, with 12 GB VRAM, and 32 GB of memory. All models were trained on Python environment, specifically with pre-trained PyTorch models from the timm library pretrained on ImageNet-1k. As we have numerous classes, The Cross-Entropy loss function was chosen because it is perfect for multi-class classification experiment.\cite{mao2023crossentropylossfunctionstheoretical}.

The models were trained for up to 100 epochs. During training, we utilized k-fold cross-validation for justification. To overcome overfitting, we employed early stopping. When the current average validation loss exceeded the best validation loss from previous epochs, we triggered the patience. If patience triggered for three consecutive epochs, training was terminated. Most models and folds converged between 90 to 100 epochs. The learning rate and batch size were set to 0.0001 and 128 respectively.

\section{Result Analysis}

In this section, we compared our model with other lightweight state-of-the-art architectures for Bengali handwritten character recognition. By evaluating different metrics such as accuracy, model size, and the number of parameters, we showed how BornoViT achieves a balance between performance and computational cost, making it an ideal solution for resource scarce environments.

\subsection{Performance of different baseline architectures}

For many years, CNN-based model has been dominating for image classification tasks. Recently, transformer based model have emerged and competing with CNN-based model. We have trained several state-of-the-art CNN and Transformer based model along with BornoViT on both Bornomala and BanglaLekha-Isolated Dataset \cite{biswas2017banglalekha} using pretrained weights.  

\begin{table}[htbp]
\caption{Performance of different baseline Architectures on our “Bornomala” Dataset. }
\centering
\begin{adjustbox}{width=\linewidth}
\begin{tabular}{l p{1.0cm} p{1.5cm} p{1.0cm} }
\hline
\textbf{Architecture} & \textbf{Accuracy (\%)} & \textbf{Trainable Parameters Count (M)} & \textbf{Model Size (MB)}  \\ \hline

MobileNetV2 \cite{sandler2019mobilenetv2invertedresidualslinear}          & \textbf{94.09}  & 2.24  & 2.13  \\ 
EfficientViT \cite{cai2024efficientvitmultiscalelinearattention}          & 92.35  & 2.14   & 2.04         \\  \hline

BornoViT (Ours)       & 91.51  & \textbf{0.65}    & \textbf{0.62}   \\ \hline

\end{tabular}
\end{adjustbox}
\label{tab:comparison_of_models_bornomala}
\end{table}

We have evaluated MobileNetV2 \cite{sandler2019mobilenetv2invertedresidualslinear} and EfficientViT \cite{cai2024efficientvitmultiscalelinearattention} on our dataset and have achieved accuracy of 94.09\% and 92.47\% respectively. Our model achieved an accuracy of 91.51\% for the Bornomala dataset, this is because our model was pretrained only on Ekush dataset \cite{ekush}, and other two models were pretrained on ImageNet-1k \cite{deng2009imagenet}.

\begin{table}[htbp]
\caption{Performance of different baseline Architectures on BanglaLekha-Isolated Dataset\cite{biswas2017banglalekha}.}
\centering
\begin{adjustbox}{width=\linewidth}
\begin{tabular}{l p{1.0cm} p{1.5cm} p{1.0cm} }
\hline
\textbf{Architecture} & \textbf{Accuracy (\%)} & \textbf{Trainable Parameters Count (M)} & \textbf{Model Size (MB)}  \\ \hline

MobileNetV2 \cite{sandler2019mobilenetv2invertedresidualslinear}          & 91.37  & 2.24  & \textbf{2.13}  \\ 
DenseNet \cite{huang2018denselyconnectedconvolutionalnetworks}          & 92.73  & 7.98   & 30.44         \\  
Xception \cite{chollet2017xceptiondeeplearningdepthwise}          & 90.34  & 20.83   & 19.86         \\  \hline

BornoViT (Ours)       & \textbf{95.77} &  \textbf{0.65}    & \textbf{0.62}   \\ \hline

\end{tabular}
\end{adjustbox}
\label{tab:comparison_of_models_banglalekha}
\end{table}

To evaluate on BanglaLekha-Isolated dataset \cite{biswas2017banglalekha}, we have utilized three different pretrained models along with BornoViT. These models are MobileNetV2 \cite{sandler2019mobilenetv2invertedresidualslinear}, DenseNet \cite{huang2018denselyconnectedconvolutionalnetworks} and Xception \cite{chollet2017xceptiondeeplearningdepthwise}. Here, BornoViT has achieved highest accuracy of 95.77\% and outperforms all other models in terms of trainable parameters count, model size.

\subsection{Class-wise Analysis}

The summarized classification report of our model performance shows that we get precision of 99\% for one character class \textit{oi} and 98\% for seven character classes.  For the class \textit{tha}, our model achieved only 81\% precision as it has inter-class similarity with character \textit{kha}, and class \textit{kha} has precision of 89\%. Our model has achieved  96\% average precision and 96\% average recall.  The character class \textit{yya} has the highest recall of 100\% and class \textit{ka} has the highest f1-score of 100\%.

\subsection{Comparison with state-of-the-art methods}

We conduct a comparison with state-of-the-art models to demonstrate, how our
lightweight model performs in the classification task of Bengali
handwritten characters. Table \ref{tab:comparison_of_performance} shows the results comparison of different models on BanglaLekha-Isolated dataset \cite{biswas2017banglalekha}. From Table \ref{tab:comparison_of_performance}, we can see that VashaNet \cite{RAQUIB2024100568} was able to achieve an accuracy of 94.78\% while having lower number of trainable parameters, only 1.04 million. The model size of VashaNet is much smaller than others. with only 2.48 GFLOPs. Although Bangla\_HCR \cite{opu2024handwritten} was able to get higher accuracy of 96.87\%, the model is heavier than others as it has 3.45 million parameters and 13.2 mb of model size. HCR-Net \cite{chauhan2024hcr} utilizes a CNN model with a higher parameter count, model size and GFLOPs, but unable to achieve the expected results. CNN-Bengali \cite{asraful2023handwritten} has lower accuracy among all other methods.

\begin{table}[htbp]
\caption{Performance comparison with the state-of-the-art
works on BanglaLekha-Isolated Dataset\cite{biswas2017banglalekha}.}
\centering
\begin{adjustbox}{width=\linewidth}
\begin{tabular}{l p{1.0cm} p{1.5cm} p{1.0cm} p{1.0cm}}
\hline
\textbf{Architecture} & \textbf{Accuracy (\%)} & \textbf{Trainable Parameters Count (M)} & \textbf{Model Size (MB)} & \textbf{FLOPs Count (GFLOPS)} \\ \hline

VashaNet \cite{RAQUIB2024100568}      & 94.78  & 1.04  & 3.95 & 2.48 \\ 
Bangla\_HCR\cite{opu2024handwritten}     & 96.87  & 3.45       & 13.2      & -     \\
HCR-Net \cite{chauhan2024hcr}     &  95.74  & 9.74    & 37.17  & 14.52  \\
CNN-Bengali \cite{asraful2023handwritten}      & 92.48  &  2.2   & 8.24  & 8.71  \\ \hline

BornoViT (Ours)       & 95.77  & \textbf{0.65}    & \textbf{0.62}   & \textbf{0.16}  \\ \hline

\end{tabular}
\end{adjustbox}
\label{tab:comparison_of_performance}
\end{table}

Our model BornoViT, performed significantly well, with higher accuracy of 95.77\%, while maintaining a lower parameter count of only 0.65 million. The model size is only 0.62 MB, with only 0.16 GFLOPs. Although our model’s accuracy is slightly lower than Bangla\_HCR \cite{opu2024handwritten} , it has far fewer parameters, smaller model size and lower GFLOps.

\subsection{Qualitative Analysis}

\begin{figure}[htbp]
    \centering
    \subfloat[Actual: \textit{gha}]{%
        \includegraphics[width=0.45\linewidth]{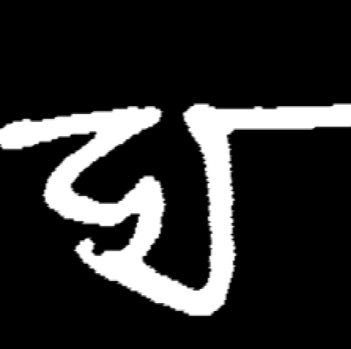}
        \label{fig:image1grad_cam}
    }\hfill
    \subfloat[Predicted: \textit{gha}]{%
        \includegraphics[width=0.45\linewidth]{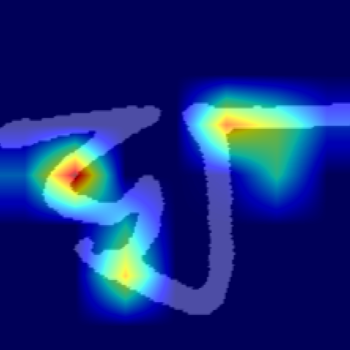}
        \label{fig:image2grad_cam}
    }\\
    \subfloat[Actual: \textit{rosho\_i}]{%
        \includegraphics[width=0.45\linewidth]{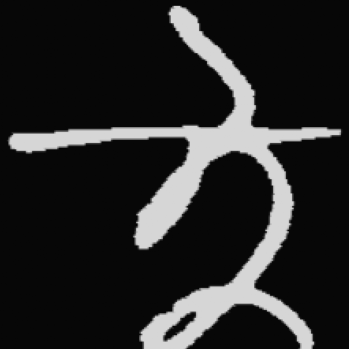}
        \label{fig:image3grad_cam}
    }\hfill
    \subfloat[Predicted: \textit{rosho\_i}]{%
        \includegraphics[width=0.45\linewidth]{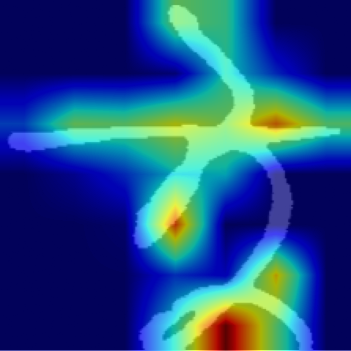}
        \label{fig:image4grad_cam}
    }
    \caption{GradCam samples of correctly classified inputs of BanglaLekha-Isolated}
    \label{fig:grad_cam}
\end{figure}

In order to figure out whether our proposed model is able to learn from the relevant spatial features and patterns, we have plotted the gradient of last attention layer for some samples using GradCAM. This attention map visualizes how perfectly our model is able to classify Bengali handwritten characters using salient features and by focusing on relevant regions of the input image. Figure \ref{fig:grad_cam} shows some attention map samples of correctly predicted inputs. This visualization shows that how good our model is to classify image, focusing on global spatial information by utilizing self-attention mechanism of transformer model.

\begin{figure}[htbp]
    \centering
    \subfloat[Actual: \textit{bha}]{%
        \includegraphics[width=0.45\linewidth]{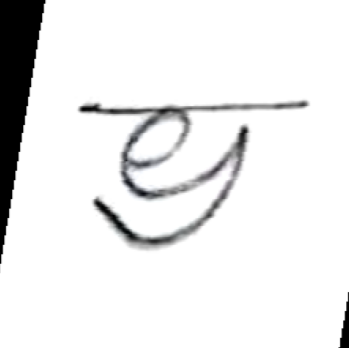}
        \label{fig:image1grad_cam_bornomala}
    }\hfill
    \subfloat[Predicted: \textit{bha}]{%
        \includegraphics[width=0.45\linewidth]{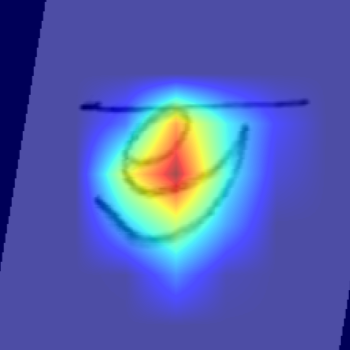}
        \label{fig:image2grad_cambornomala}
    }\\
    \subfloat[Actual: \textit{pach}]{%
        \includegraphics[width=0.45\linewidth]{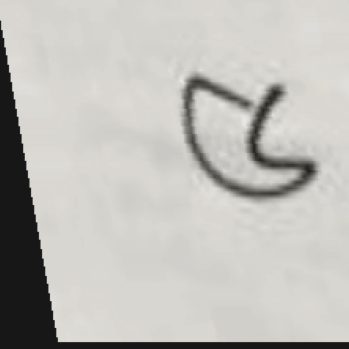}
        \label{fig:image3grad_cambornomala}
    }\hfill
    \subfloat[Predicted: \textit{pach}]{%
        \includegraphics[width=0.45\linewidth]{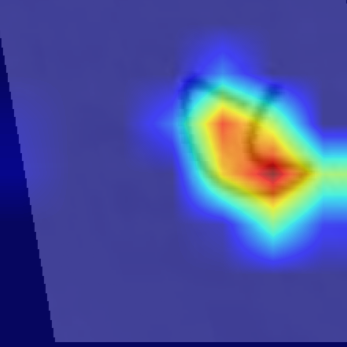}
        \label{fig:image4grad_cambornomala}
    }
    \caption{GradCam samples of correctly classified inputs of Bornomala}
    \label{fig:grad_cam_bornomala}
\end{figure}

Figure \ref{fig:grad_cam_bornomala} shows some examples of correctly predicted output of BornoViT model of Bornomala Dataset. From figure \ref{fig:image2grad_cambornomala}, we can see that our model is able to focus of spatial feature and can utilize the character pattern and area effectively. For Bengali digit \textit{pach}, BornoViT is exactly focusing on unique pattern and feature of \textit{pach}, and predicted the class utilizing salient features.

\begin{figure}[htbp]
    \centering
    \subfloat[Predicted: \textit{jha} Actual: \textit{ri}]{%
        \includegraphics[width=0.45\linewidth]{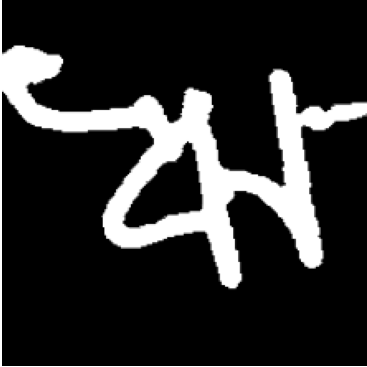}
        \label{fig:image1_misclass}
    }\hfill
    \subfloat[Predicted: \textit{ta} Actual: \textit{uo}]{%
        \includegraphics[width=0.45\linewidth]{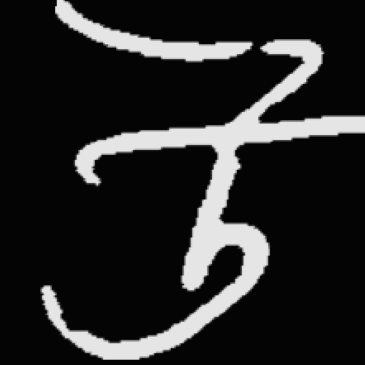}
        \label{fig:image2_misclass}
    }\\
    \subfloat[Predicted: \textit{tha} Actual: \textit{kha}]{%
        \includegraphics[width=0.45\linewidth]{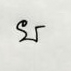}
        \label{fig:image3_misclass}
    }\hfill
    \subfloat[Predicted: \textit{kha} Actual: \textit{tha}]{%
        \includegraphics[width=0.45\linewidth]{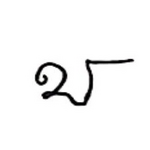}
        \label{fig:image4_misclass}
    }
    \caption{Misclassified samples that are visually similar to the predicted class.}
    \label{fig:missclassified_fig}
\end{figure}

However, for some input images, our model failed to make correct prediction. The principal cause behind these misclassifications is inter-class similarity. Figures \ref{fig:image3_misclass} \& \ref{fig:image4_misclass} show that both the character \textit{kha} and \textit{tha} have similar patterns. Figure \ref{fig:image1_misclass} is a sample of class \textit{rii} but model is predicting as of class \textit{jha}. Similar goes for the figure \ref{fig:image2_misclass}. As a result, for some input images, our model makes incorrect prediction. There are numerous classes that have inter-class similarity. Also, the writing style of individuals can vary for the same character, causing intra-class dissimilarity, because of this, our model sometimes fails to predict these characters accurately.

\section{Conclusion \& Future Work }

We conducted experiment to solve classification task of Bengali handwritten basic characters and digits, using our own Vision Transformer based model BornoViT. We achieved a dramatically decreased computational cost with no significant loss of classification performance. Despite being a low capacity model with 0.65 million parameters, BornoViT is able to achieve competitive results. The results also show the capability of BornoViT to overcome the drawbacks of contemporary approaches, which demand huge computational power and large data, thus making it a viable alternative for deployment in resource-limited settings.

We plan to continue this in the future and do further research along several lines to improve BornoViT. We will refine our model further based on more complex characters with larger datasets from different sources possible. We also plan to augment the data and apply transfer learning such that the model can generalize for a range of handwriting styles, as well as extend BornoViT to other low-resource languages.

\bibliographystyle{ieeetr}
\bibliography{citations} % this is using to cite or reference purpose

\end{document}